\documentclass{article}
\usepackage[utf8]{inputenc}
\usepackage{color}
\usepackage{multirow}
\usepackage{amsmath}
\usepackage{amssymb}
\usepackage{stmaryrd}
\usepackage{graphicx}
\usepackage[numbers,square]{natbib}
\usepackage{microtype}

\makeatletter

\providecommand{\tabularnewline}{\\}

\usepackage[preprint]{neurips_2018}

\usepackage{times}
\usepackage{soul}
\usepackage[small]{caption}
\usepackage{amsthm}
\usepackage{algorithm}
\usepackage{algorithmic}

\title{IJCAI--PRICAI--20 Formatting Instructions}

\author{Anonymous Submission
}

\author{%
  Thao Minh Le, Vuong Le, Svetha Venkatesh, Truyen Tran \\
  Applied Artificial Intelligence Institute, Deakin University, Australia\\
  \texttt{\{lethao,vuong.le,svetha.venkatesh,truyen.tran\}@deakin.edu.au} \\
}

\makeatother

\begin{document}
\title{Dynamic Language Binding in Relational Visual Reasoning}

\maketitle
\global\long\def\ModelName{\text{LOGNet}}%
\global\long\def\UnitName{\text{LOG}}%
\global\long\def\Problem{\text{VQA}}%
\global\long\def\BindingConstructor{\text{Language binding constructor}}%
\global\long\def\GraphConstructor{\text{Visual graph constructor}}%
\global\long\def\Refinement{\text{Representation Refinement}}%

\begin{abstract}
We present Language-binding Object Graph Network, the first neural
reasoning method with dynamic relational structures across both visual
and textual domains with applications in visual question answering.
Relaxing the common assumption made by current models that the object
predicates pre-exist and stay static, passive to the reasoning process,
we propose that these dynamic predicates expand across the domain
borders to include pair-wise visual-linguistic object binding. In
our method, these contextualized object links are actively found within
each recurrent reasoning step without relying on external predicative
priors. These dynamic structures reflect the conditional dual-domain
object dependency given the evolving context of the reasoning through
co-attention. Such discovered dynamic graphs facilitate multi-step
knowledge combination and refinements that iteratively deduce the
compact representation of the final answer. The effectiveness of this
model is demonstrated on image question answering demonstrating favorable
performance on major VQA datasets. Our method outperforms other methods
in sophisticated question-answering tasks wherein multiple object
relations are involved. The graph structure effectively assists the
progress of training, and therefore the network learns efficiently
compared to other reasoning models. 

\end{abstract}

\section{Introduction}

Reasoning is crucial for intelligent agents wherein relevant clues
from a knowledge source are retrieved and combined to solve a query,
such as answering questions about an image. Human visual reasoning
involves analyzing linguistic aspects of the query and continuously
inter-linking them with visual objects through a series of information
aggregation steps \citep{lake2017building}. Artificial reasoning
engines mimic this ability by using structured representations (e.g.
scene graphs) \citep{shi2019explainable} to discover categorical
and relational information about visual objects.

In this work, we address two key abstractions: How can we extend this
structure seamlessly across both visual-lingual borders? And, unlike
prior work, how can we extend these structures to be dynamic and responsive
to the reasoning process? We explore the dynamic relational structures
of visual scenes that are proactively discovered within reasoning
context and their adaptive connections to the components of a linguistic
query to effectively answer visual questions. 

\begin{figure}
\begin{centering}
\includegraphics[width=0.8\columnwidth]{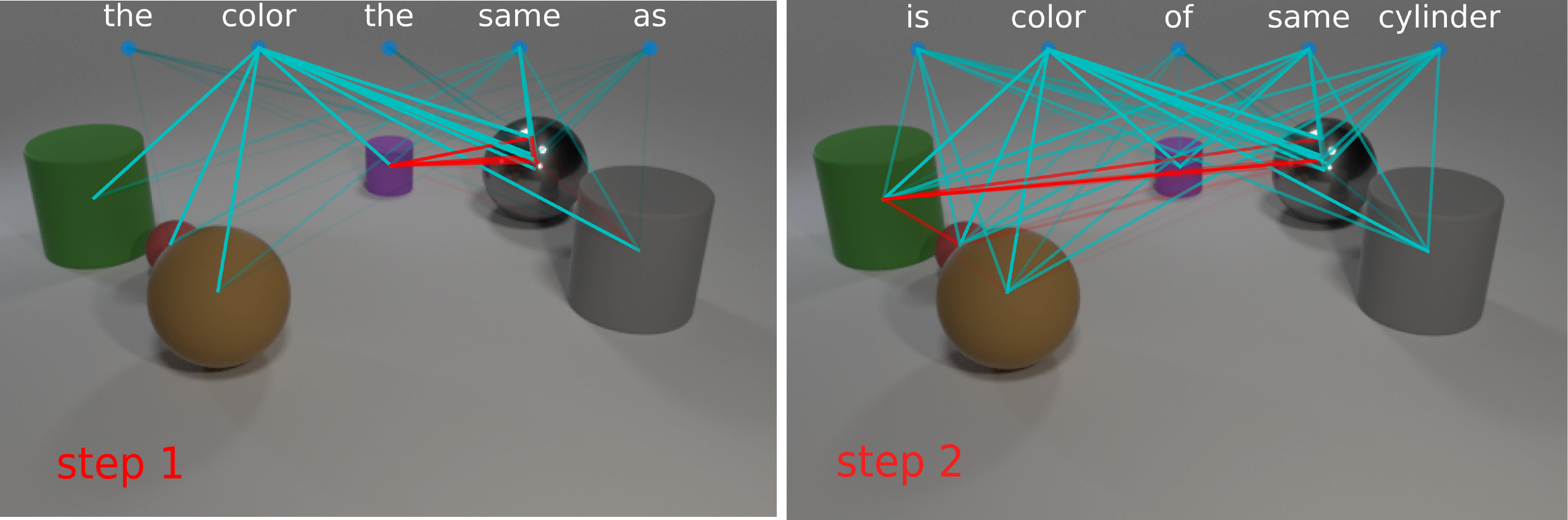}
\par\end{centering}
\caption{We aim to dynamically construct visual graphs (red edges) and linguistic-visual
bindings (cyan edges (most prominent words shown)) adaptively to reasoning
steps for each image-question pair. \label{fig:illustration}}
\end{figure}

Recent history observes the success of compositional reasoning which
iteratively pays attention to a subset of clues in the query and simultaneously
looks up a corresponding subset of facts from a static unstructured
knowledge source to construct a representation related to the answer
\citep{hudson2018compositional}. Concurrently, findings in visual
relational modeling show that the information in visual scenes is
significantly distributed at the interconnections between semantic
factors of visual objects and linguistic objects from both the image
and query \citep{baradel2018object}. These observations suggest that
relational structures can improve compositional reasoning \citep{xu2019can}.
However, direct application of attention mechanisms on a static structuralized
knowledge source \citep{velivckovic2018graph} would miss the full
advantage of compositionality. Moreover, object relations are naturally
rich and multifaceted \citep{kim2018visual}, therefore an \emph{a
priori} defined set of semantic predicates such as visual scene graphs
\citep{hudson2019learning} and language grounding \citep{huang2019multi}
are either incomplete \citep{xu2017scene}, or too complicated and
irrelevant to use without further pruning. 

We approach this dilemma by dynamically constructing relevant object
connections on-demand according to the evolving reasoning states.
There are two types of connections: links that relate visual objects
and links that bind visual objects in the image to linguistic counterparts
in the query (See Fig. \ref{fig:illustration}). Conceptually, this
dynamic structure constitutes a relational working memory that temporarily
links and refines concepts both within and across modalities. These
relations are compact and readily support structural inference.

Our model, called Language-binding Object Graph Networks ($\ModelName$)
for visual question answering ($\Problem)$, includes an iterative
operation of LOG unit that uses a contextualized co-attention to identify
pairs of visual objects that are temporally related. Another co-attention
head is concurrently used to provide cross-domain binding between
visual concepts and linguistic clues. A progressive chain of dynamic
graphs is inferred by our model (see Fig.~\ref{fig:illustration}).
These dynamic structures enable representation refinement with residual
graph convolution iterations. The refined information will be added
to an internal working memory progressing toward predicting the answer.
The modules are interconnected through co-attention signals making
the model end-to-end differentiable.

We apply our model on major $\Problem$ datasets. Both qualitative
and quantitative results indicate that $\ModelName$ has advantages
over state-of-the-art methods in answering long and complex questions.
Our results show superior performance even when trained on just 10\%
of data. These questions require complex high-order reasoning which
necessitates our model's ability to dynamically couple entities to
build a predicate, and then chain these predicates in the correct
order. The structured representation provides guidance to the reasoning
process, improving the fitness of the learning particularly with limited
training data.

\section{Related Work}

Recent compositional reasoning research aims at either structured
symbolic program execution using custom built modules \citep{hu2017learning}
or working through recurrent implicit reasoning steps on an unstructured
representation \citep{perez2018film}. Relational structures have
been demonstrated to be crucial for reasoning \citep{xu2019can}.
End-to-end relational modeling considers pair-wise predicates of CNN
features \citep{santoro2017simple}. With reliable object detection,
visual reasoning can use semantic objects as cleaner representations
\citep{anderson2018bottom,desta2018object}. When semantic or geometrical
predicate labels are available, either as provided \citep{hudson2019gqa}
or by learning \citep{xu2017scene} to form semantic scene graphs,
such structures can be leveraged for visual reasoning \citep{shi2019explainable,li2019relation}.
In contrast to these methods, our relational graphs are not limited
by the predefined predicates but liberally form them according to
the reasoning context. Our model is also different from previous question-conditioned
graph construction \citep{norcliffe2018learning} in the dynamic nature
of the multiform graphs where only relations that are relevant emerge.
Dynamic graph modeling has been considered by recurrent modeling \citep{palm2018recurrent},
and although their states transform, the graph structures stay fixed.
A related idea uses language conditioned message passing to extract
context-aware features \citep{hu2019language}. In contrast, $\ModelName$
does not treat linguistic cues as a single conditioning vector, but
allows them to live as a set of active objects that interact with
visual objects through binding and individually contribute to the
joint representation. The language binding also differentiates $\ModelName$
from MUREL \citep{cadene2019murel} where the contributions of linguistic
cues to visual objects are the same though an expensive bilinear operator.

\section{Language-binding Object Graph Network}

The goal of a VQA task is to deduce an answer $\tilde{a}$ from an
image $I$ in response to a natural question $q$. Let the answer
space be $\mathbb{A}$, VQA is formulated as: 
\begin{equation}
\tilde{a}=\arg\max_{a\in\mathbb{A}}\mathcal{P}_{\theta}\left(a\mid q,I\right),
\end{equation}
where, $\theta$ is the learnable parameters of $\mathcal{P}$.

\begin{figure*}
\begin{centering}
\includegraphics[width=0.95\textwidth]{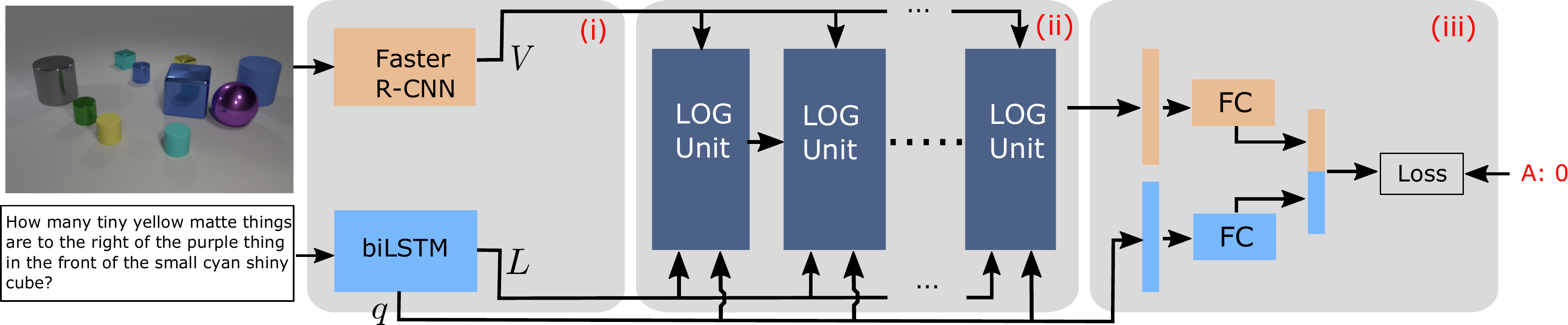}
\par\end{centering}
\caption{Overall Architecture of $\protect\ModelName$. (i) Linguistic and
visual representations (ii) Information refinement with $\protect\UnitName$
modules (iii) Multimodal fusion and answer prediction.\label{fig:Overall-Architecture}}
\end{figure*}

We envision VQA as a process of relational reasoning over a scene
of multiple visual objects conditioned on a set of linguistic cueing
objects. Crucially, a pair of co-appearing visual objects may induce
multiple relations, whose nature may be unknown \emph{a priori}, and
hence must be inferred dynamically in adaptive interaction with the
linguistic cues.

We present a new neural model $\mathcal{P}$ called $\ModelName$
(See Fig.~\ref{fig:Overall-Architecture}) to realize this vision.
At the high level, for each image and query pair, $\ModelName$ first
normalizes them into two individual sets of linguistic and visual
objects. Then, it performs iterative multi-step reasoning by iteratively
summoning Language-binding Object Graph ($\UnitName$) units to achieve
a compact multi-modal representation in a recurrent manner. This representation
is finally combined with the query representation to reach the answers.
We detail these steps.

\subsection{Linguistic and Visual Objects}

We embed words in the length-$S$ query into 300-D vectors, which
are subsequently passed through a biLSTM. The hidden states of LSTM
representing the context-dependent word embeddings $e_{s}$ are collected
into a chain of contextual embeddings $L=\left\{ e_{s}\right\} _{s=1}^{S}$
$\in\mathbb{R}^{d\times S}$ and used as linguistic objects in reasoning.
We also retain the overall query semantic as $q=\left[\overleftarrow{e_{1}};\overrightarrow{e_{S}}\right]$
which joins the final states of forward and backward LSTM passes.
Unless otherwise specified, we use $[.\thinspace;.]$ to denote the
concatenation operator of two tensors.

The input image $I$ is first processed into a set of appearance/spatial
features $O=\left\{ (a_{i},p_{i})\right\} _{i=1}^{N}$ of $N$ regions
extracted by an off-the-shelf object detection such as Faster R-CNN
\citep{ren2015faster}. The appearance component $a_{i}\in\mathbb{R^{\text{2048}}}$
are ROI pooling features and the spatial $p_{i}$ are normalized coordinates
of the region box \citep{yu2017joint}. These features are further
combined and projected by trainable linear embeddings to produce a
set of visual objects $V=\{v_{i}\}_{i=1}^{N}\in\mathbb{R}^{d\times N}$.
The pair $(L,V)$ are readily used as input for a chain of $\UnitName$
reasoning operations.

\subsection{Language-binding Object Graph Unit}

$\UnitName$ is essentially a recurrent unit whose state is kept in
a compact working memory $m_{t}$ and a controlling signal $c_{t}$.
Input of each $\UnitName$ operation includes the visual and linguistic
objects $(V,L)$, and the overall query semantic $q$.

\begin{figure*}
\begin{centering}
\includegraphics[width=1\textwidth]{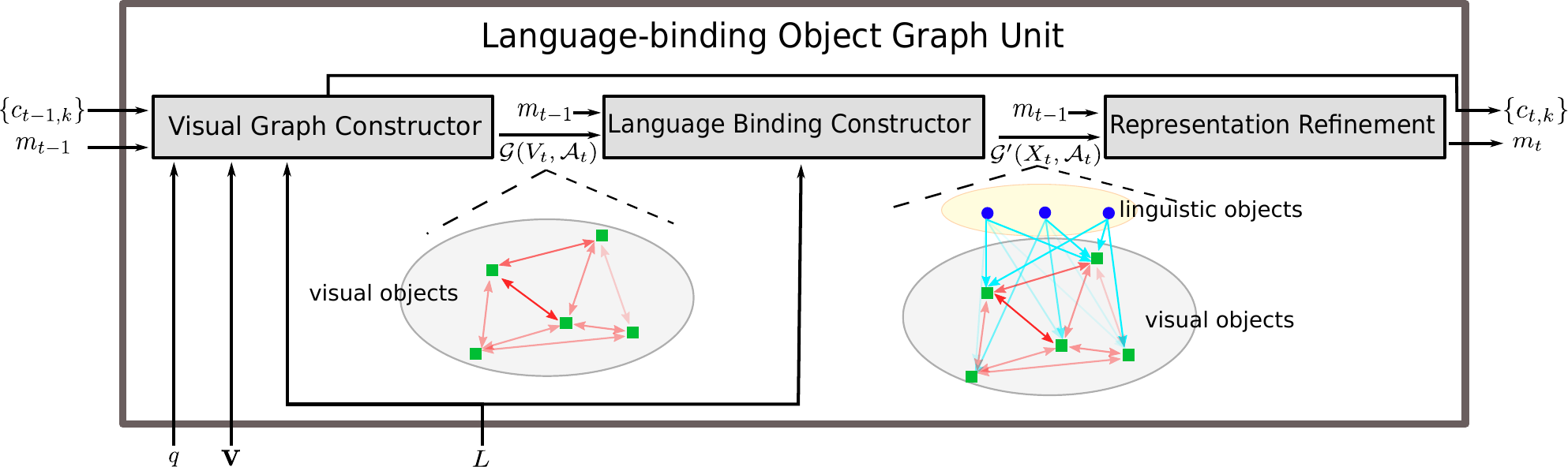}
\par\end{centering}
\caption{Language-binding Object Graph (LOG) Unit. $L$: linguistic objects,
$V$: visual objects, red edges: visual graph, cyan edges: language-visual
binding. The following elements are dynamic at pass $t$: $q_{t}$
-- query semantic ; $\{c_{t,k}\}$ -- language-based controlling
signals; $m_{t}$ - working memory state. \label{fig:Graph-Structured-Co-Attention-Mo}}
\end{figure*}

Each $\UnitName$ consists of three submodules: (i) a \emph{visual
graph constructor }to build a context-aware weighted adjacency matrix
of visual graph $\mathcal{G}_{t}$, (ii) a \emph{language binding
constructor }to compute the adaptive linkage between linguistic and
visual objects and form a multi-modal graph $\mathcal{G}_{t}^{\prime}$
(iii)\emph{ representation refinement }module\emph{ }to update object
representation using the graphs. (See Fig. \ref{fig:Graph-Structured-Co-Attention-Mo}).

\subsubsection{$\protect\GraphConstructor$}

At each $\UnitName$ operation, we construct an undirected graph $\mathcal{G}_{t}=\left(\mathit{\mathit{V}}_{t},\mathcal{A}_{t}\right)$
from $N$ visual objects $V=\{v_{i}\}_{i=1}^{N}$ by finding adaptive
features $V_{t}$ and constructing the weighted adjacency matrix $\mathcal{A}_{t}$.
Different from the widely used static semantic graphs \citep{xu2017scene},
our graph $\mathcal{G}_{t}$ is dynamically constructed at each reasoning
step $t^{th}$ and is modulated by the recurrent controlling signal
$c_{t}$ and overall linguistic cue $q$. This reflects the dynamic
relations of objects triggered by both the question and reasoning
context. In fact, this design is consistent with how human reasons.
For example, looking at an image, to answer different questions, we
connect different pairs of objects although their geometrical and
appearance similarities were unchanged. Moreover, even at one question,
our mind traverses through multiple types of object relationships
in different steps of reasoning, especially when a query contains
multiple or nested relations. Let $W_{t}$ denote sub-networks' weights
at step $t^{th}$, we first augment the nodes' features as
\begin{equation}
V_{t}=W_{t}^{v}\left[V;m_{t-1}\varodot V\right]+b^{v}.\label{eq:6}
\end{equation}

The controlling signals $\{c_{t,k}\}$ is derived from its previous
state and a step-specific query semantic $q_{t}$ through a set of
$K$ attention heads $\{\alpha_{t,k}\}_{k=1}^{K}$ on the linguistic
objects $L=\left\{ e_{s}\right\} _{s=1}^{S}$:
\begin{align}
c_{1} & =q_{1},\qquad q_{t}=W_{t}^{q}q+b_{t}^{q}\\
q_{t}^{\prime} & =[q_{t};\sum_{k=1}^{K}(\gamma_{t,k}*c_{t-1,k})],\qquad\sum_{k=1}^{K}\gamma_{t,k}=1,\\
\alpha_{s,t,k} & =\text{softmax}_{s}\left(W_{t,k}^{\alpha}(e_{s}\varodot q_{t}^{\prime})\right),\\
c_{t,k} & =\sum_{s=1}^{S}\alpha_{s,t,k}*e_{s},\,\,\,c_{t}=\{c_{t,k}\},
\end{align}
where, $\gamma_{t,k}$ is the weights of the past controlling signals
being added to the current query semantic $q_{t}^{\prime}$.

While single attention can be used to guide the multi-step reasoning
process \citep{hudson2018compositional}, we noticed that it tends
to focus on one object attribute at a time neglecting inter-aspect
relations because of the softmax operation. In $\Problem$, multiple
object attributes are usually necessary - e.g. to answer ``what is
the color of the small shiny object having the same shape with the
cyan sphere?'', the object aspects ``color'' and ``shape'' both
need to be attended to. Our development of using multi-head attention
enables such a goal. The controlling signals are then used to build
the context modulated node description matrix of $r$ rows, $\tilde{V}_{t}\in\mathbb{R}^{r\times N}$
:

\begin{equation}
\tilde{V}_{t}=\textrm{norm}\left(W_{t}^{\tilde{v}}\sum_{k=1}^{K}(V\varodot c_{t,k})\right),
\end{equation}
where, $norm$ is a normalization function for numerical stabilization
which is softmax in our implementation.

Finally, we estimate the symmetric adjacency matrix $\mathcal{A}_{t}$$\in\mathbb{R}^{N\times N}$
by relating node features in $\tilde{V_{t}}$. $\mathcal{A}_{t}$
is a rank $r$ symmetric matrix representing the first-order proximity
in appearance and spatial features of the nodes:
\begin{equation}
\mathcal{A}_{t}=\tilde{V}_{t}^{\top}\tilde{V}_{t}.
\end{equation}

The motivation behind the estimation of $\mathcal{A}_{t}$ is similar
to recent works \citep{santoro2017simple,cadene2019murel} on modeling
\emph{implicit} relations of visual objects, in which they do not
reflect any semantic or spatial relations but indicate the probabilities
of object-pair co-occurrences given a query.

\subsubsection{$\protect\BindingConstructor$}

The visual graph explored by the visual graph constructor is powerful
in representing dynamic object relation albeit still lacking the two-way
complementary object-level relation between visual and textual data.
In one direction, visual features provide grounding to ambiguous linguistic
words so that objects of the same category can be differentiated \citep{nagaraja2016modeling}.
Imagine the question ``what is the color of the cat eating the cake''
in a scene with many cats visible, then appearance and spatial features
will clarify the selection of the cat of interest. In the opposite
direction, linguistic cues provide more precise information than visual
features of segmented regions. In the previous example, the ``eat''
relation between ``cat'' and ``cake'' is clear from the query
words and is useful to connect these two visual objects in the image.
These predicative advantages are even more important in the case of
higher order relationships.

Drawing inspiration from that observation, we build a multi-modal
graph $\mathcal{G}_{t}^{\prime}=(X_{t},\mathcal{A}_{t})$ from the
constructed graph $\mathcal{G}_{t}=(\mathcal{\mathit{V}}_{t},\mathcal{A}_{t})$.
Each node $x_{t,i}\in X_{t}$ of $\mathcal{G}_{t}^{\prime}$ is a
binding of the corresponding visual node $v_{t,i}$ of $\mathcal{G}_{t}$
with its linguistic supplement given by the context-aware function
$f_{t}(.)$:

\begin{equation}
x_{t,i}=[v_{t,i};f_{t}(e_{1},...,e_{S}|v_{t,i})].
\end{equation}

Designing $f_{t}(.)$ is key to make this representation meaningful.
In particular, we design this function as the weighted composition
of contextual words $\{e_{s}\}_{s=1}^{S}$:
\begin{equation}
f_{t}(e_{1},...,e_{S}|v_{t,i})=\sum_{s=1}^{S}\beta_{t,i,s}*e_{s}.
\end{equation}
Here combination weights $\beta_{t,i,s}$ represent the cross-modality
partnership between a visual object $v_{t,i}$ and a linguistic word
$e_{s}$, essentially forming the contextualized pair-wise bipartite
relations between the $V$ and $L$. 

To calculate $\beta_{t,i,s}$, we first preprocess them by modulating
$V$ with the previous memory state $\hat{V}_{t}=W_{t}^{\hat{v}}\left[V;m_{t-1}\varodot V\right]+b^{\hat{v}}$
and soft classifying each word $s$ into multiple lexical types as
a weight vector $z_{s}$ similar to \citep{yang2019dynamic}, $z_{s}=\sigma(W^{z1}(W^{z0}e_{s}+b^{z0})+b^{z1}).$
Subsequently, the normalized cross-modality relation weights are calculated
as:

\begin{align}
\beta_{t,i,s}= & z_{s}*\textrm{softmax}_{s}(W_{t}^{\beta}(\text{tanh}(W_{t}^{\hat{v}}\hat{v}_{t,i}+W_{t}^{e}e_{s}))).
\end{align}

By doing this, we allow per-object communication between the two modalities,
differentiating our method from prior works where linguistic cue is
reduced to a single vector for conditioning or combined with visual
signal in a late fusion.

\subsubsection{$\protect\Refinement$}

At the last step of $\UnitName$ operation, we rely on the newly built
multi-modal graph $\mathcal{G}_{t}^{\prime}=(X_{t},\mathcal{A}_{t})$
as the structure to refine the representation of objects by employing
a graph convolutional network (GCN) \citep{kipf2016semi} of $H$
hidden layers. Generally, vanilla GCNs have a difficulty of stacking
deep layers due to the common vanishing gradient and numerical instability.
We solve this problem by borrowing the residual skip-connection trick
from ResNet \citep{he2016deep} to create more direct gradient flow.
Concretely, the refined node representation is given by:

\begin{align}
R_{1} & =X_{t},\\
F_{h}\left(R_{h-1}\right) & =W_{h-1}^{2}\rho\left(W_{h-1}^{1}R_{h-1}\mathcal{A}_{t}+b_{h-1}\right),\\
R_{h} & =\rho\left(R_{h-1}+F_{h}\left(R_{h-1}\right)\right),
\end{align}
where, $h=1,2..,H$, and $\rho$ is an activation function which is
an ELU operation in our later experiments. The parameters $\left(W_{h-1}^{1},W_{h-1}^{2}\right)$
can be optionally tied across $H$ layers.

As we obtain the refined representation $R_{t,H}=\{r_{t,i,H}\}_{i=1}^{N}$
after the $H$ refinement layers, we compute the overall final representation
by smashing the graph into one single vector:

\begin{equation}
\tilde{x_{t}}=\sum_{i=0}^{N}\delta_{t,i}*r_{t,i,H},
\end{equation}
where, $\delta_{t,i}=\text{softmax}_{i}(W_{t}^{\delta}r_{t,i,H})$.
Finally, we update $\UnitName$'s working memory state:

\begin{equation}
m_{t}=W_{t}^{m}\left[m_{t-1};\tilde{x}_{t}\right]+b^{m}.
\end{equation}

\subsection{Answer Prediction}

After $T$ passes of $\UnitName$ iterations, $\ModelName$ combines
the final memory state $m_{T}$ with the sequential expression $q$
of the question by concatenation followed by a linear layer to get
the final representation $J=W\left[m_{T};q\right]+b,\ J\in\mathbb{R^{\text{d}}}$.

For answer prediction, we adopt a 2-layer multi-layer perceptron (MLP)
and a batch normalization layer in between as a classifier. The network
is trained using cross-entropy loss or binary cross-entropy loss according
to types of questions.

\section{Experiments}

\begin{figure}
\noindent\begin{minipage}[c]{1\textwidth}%
\begin{minipage}[c]{0.55\textwidth}%
\centering \includegraphics[width=1\columnwidth]{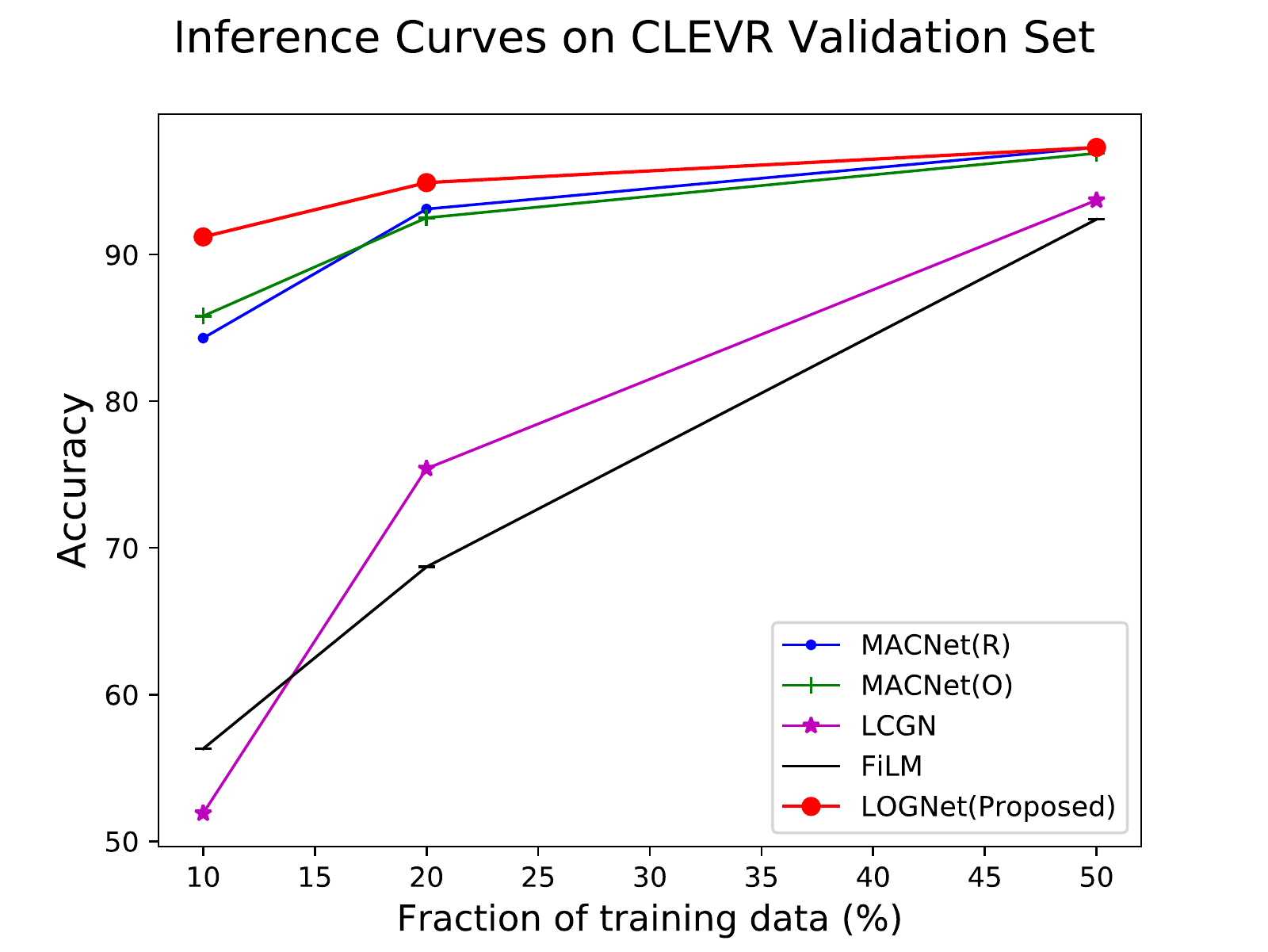}
\captionof{figure}{VQA performance on CLEVR subsets.\label{fig:Model-performance-on}}%
\end{minipage}\hfill{}%
\begin{minipage}[c]{0.45\textwidth}%
\centering %
\begin{tabular}{l|c}
\hline 
\multirow{1}{*}{{Method}} & \multicolumn{1}{c}{{Val. Acc. (\%)}}\tabularnewline
\hline 
{FiLM} & {56.6}\tabularnewline
{MACNet(R)} & {57.4}\tabularnewline
{LCGN \citet{hu2019language}} & {46.3}\tabularnewline
{BAN \citet{shrestha2019answer}} & {60.2}\tabularnewline
{RAMEN \citet{shrestha2019answer}} & {57.9}\tabularnewline
\textbf{LOGNet} & \textbf{62.5}\tabularnewline
\hline 
\end{tabular}\smallskip{}
 \captionof{table}{Performance on CLEVR-Human. \label{tab:clevr-human}}%
\end{minipage}%
\end{minipage}
\end{figure}

\subsection{Datasets}

We evaluate our model on multiple datasets including:

\textbf{CLEVR} \citet{johnson2017clevr}: presents several reasoning
tasks such as transitive relations and attribute comparison. We intentionally
design experiments to evaluate the generalization capability of our
model on various subsets of CLEVR, where most existing works fail,
sampled by the number of questions.

\textbf{CLEVR-Human }\citet{johnson2017inferring}:\textbf{ }composes
natural language question-answer pairs on images from CLEVR. Due to
diverse linguistic variations, this dataset requires stronger visual
reasoning ability than CLEVR.

\textbf{GQA} \citet{hudson2019gqa}:\textbf{ }the current largest
visual relational reasoning dataset providing semantic scene graphs
coupled with images. Because $\ModelName$ does not need prior predicates,
we ignore these static graphs using only the image and textual query
as input.

\textbf{VQA v2 }\citet{goyal2017making}: As a large portion of questions
is short and can be answered by looking for facts in images, we design
experiments with a split of only long questions (\textgreater 7 words).
The split, hence, assesses the ability to model the relations between
objects, e.g.: ``What is the white substance on the left side of
the plate and on top of the cake?''.

\subsection{Performance Against SOTAs}

\begin{table*}
\begin{minipage}[c][1\totalheight][b]{0.45\columnwidth}%
\begin{center}
\begin{tabular}{ll|cc}
\hline 
\multirow{2}{*}{Training size} & \multirow{2}{*}{Method} & \multicolumn{2}{c}{Accuracy (\%)}\tabularnewline
\cline{3-4} \cline{4-4} 
 &  & val & test\tabularnewline
\hline 
\multirow{5}{*}{\textbf{Full}} & CNN+LSTM & 49.2 & 46.6\tabularnewline
 & Bottom-Up & 52.2 & 49.7\tabularnewline
 & MACNet(O) & 57.5 & 54.1\tabularnewline
 & LCGN & 63.9 & 56.1\tabularnewline
 & LOGNet & 63.2 & 55.2\tabularnewline
\hline 
\multirow{2}{*}{\textbf{50\%}} & LCGN & 60.6 & -\tabularnewline
 & LOGNet & 61.0 & -\tabularnewline
\hline 
\multirow{2}{*}{\textbf{20\%}} & LCGN & 53.2 & -\tabularnewline
 & LOGNet & 53.8 & -\tabularnewline
\hline 
\end{tabular}
\par\end{center}
\caption{Performance on GQA and subsets. \label{tab:GQA}}
\end{minipage}\hspace{3em}%
\begin{minipage}[c][1\totalheight][b]{0.45\columnwidth}%
\begin{center}
\begin{tabular}{l|c}
\hline 
\multirow{1}{*}{Method} & \multirow{1}{*}{Val. Acc. (\%)}\tabularnewline
\hline 
XNM & 43.4\tabularnewline
MACNet(R) & 40.7\tabularnewline
MACNet(O) & 45.5\tabularnewline
\textbf{LOGNet} & \textbf{46.8}\tabularnewline
\hline 
\end{tabular}
\par\end{center}
\caption{Experiments on VQA v2 subset of long questions. \label{tab:vqa}}
\end{minipage}
\end{table*}

Our model is generally implemented with feature dimension $d=512$,
reasoning depth $T=8$, GCN depth $H=8$ and attention-width $K=2$.
The number of regions is $N=14$ for CLEVR and CLEVR-Human, and 100
for GQA and 36 for VQA v2 to match with other related methods. We
also match the word embeddings with others by using random vectors
of a uniform distribution for CLEVR/CLEVR-Human and pretrained GloVe
vectors for the other datasets. Pytorch implementation of our model
is available online\footnote{https://github.com/thaolmk54/LOGNet-VQA}.

We compare with state-of-the-art methods reporting performance as
in their papers or obtained with their public code. For the better
judgement of whether the improvement is from the model designs or
from the use of better visual embeddings, we reimplement MACNet \citet{hudson2018compositional}
with their feature choice of ResNet - MACNet(R), and additionally
try it out on our ROI pooling features - MACNet(O).

\subsubsection{CLEVR and CLEVR-Human Dataset}

Fig.~\ref{fig:Model-performance-on} demonstrates the large improvement
of $\ModelName$ over SOTAs including MACNet, FiLM and LGCN particularly
with limited training data. With enough data, all models converge
in performance. With smaller training data, other methods struggle
to generalize, while $\ModelName$ maintains stable performance. With
10\% of training data, FiLM quickly drops to 51.9\%, and only 48.9\%
in case of LGCN, which barely surpasses the linguistic bias performance
of 42.1\% reported by \citet{johnson2017clevr}. Behind $\ModelName$
(91.2\%), MACNet is the runner up in generalization with around 85.8\%.

Our model shows significant improvement over other works on CLEVR-Human
dataset (See Table \ref{tab:clevr-human}) where language vocab is
richer than the original CLEVR. We only report results without fine-tune
on CLEVR for better judgment of the generalization ability. This suggests
that $\ModelName$ can better handle the linguistic variations by
its advantage in modeling cross-modality interactions.

\subsubsection{GQA}

$\ModelName$ outperforms previous works including simple fusion approaches
CNN+LSTM and Bottom-Up \citet{anderson2018bottom}{\small{}{}, }and
the recent advanced multi-step inference MACNet. Although $\ModelName$
achieves competitive performance as compared with LCGN on the full
set, it shows its advantage in generalization and robustness against
overfitting in limited data experiments (20\% and 50\% splits) - see
Table~\ref{tab:GQA}.

\subsubsection{VQA v2 - Subset of Long Questions}

$\ModelName$ is finally applied to the most difficult questions of
VQA v2. Empirical results show that our model achieves favorable performance
over MACNet and XNM {\small{}{}\citet{shi2019explainable}} on this
subset. Due to the rich language vocab of human annotated datasets,
the improvements are less noticeable as compared with those on synthetic
datasets such as CLEVR.

\begin{table}[t]
\begin{centering}
{\footnotesize{}{}}%
\begin{tabular}{cl|c}
\hline 
{\small{}{}No.}  & {\small{}{}Model}  & {\small{}{}Val. Acc. (\%)}\tabularnewline
\hline 
{\small{}{}1}  & {\small{}{}Default config. (8 $\UnitName$ units, 8 GCNs)}  & {\small{}{}91.2}\tabularnewline
{\small{}{}2}  & {\small{}{}w/o bounding box features}  & {\small{}{}86.5}\tabularnewline
{\small{}{}3}  & {\small{}{}Graph constructor w/o previous memory}  & {\small{}{}86.5}\tabularnewline
{\small{}{}4}  & {\small{}{}Graph constructor w/o language}  & {\small{}{}56.2}\tabularnewline
{\small{}{}5}  & {\small{}{}Single-head attn. controlling signal}  & {\small{}{}86.3}\tabularnewline
{\small{}{}6}  & {\small{}{}Rep. refinement w/ 1 GCN layers}  & {\small{}{}75.9}\tabularnewline
{\small{}{}7}  & {\small{}{}Rep. refinement w/ 4 GCN layers}  & {\small{}{}89.4}\tabularnewline
{\small{}{}8}  & {\small{}{}Rep. refinement w/ 12 GCN layers}  & {\small{}{}91.1}\tabularnewline
{\small{}{}9}  & {\small{}{}Rep. refinement w/ 16 GCN layers}  & {\small{}{}89.5}\tabularnewline
{\small{}{}10}  & {\small{}{}Language binding w/o previous memory}  & {\small{}{}90.8}\tabularnewline
{\small{}{}11}  & {\small{}{}w/o language binding}  & {\small{}{}89.9}\tabularnewline
{\small{}{}12}  & {\small{}{}1 $\UnitName$ unit}  & {\small{}{}69.0}\tabularnewline
{\small{}{}13}  & {\small{}{}4 $\UnitName$ units}  & {\small{}{}76.3}\tabularnewline
{\small{}{}14}  & {\small{}{}12 $\UnitName$ units}  & {\small{}{}91.6}\tabularnewline
{\small{}{}15}  & {\small{}{}16 $\UnitName$ units}  & {\small{}{}91.1}\tabularnewline
\hline 
\end{tabular}\medskip{}
\par\end{centering}
\caption{Ablation studies - CLEVR dataset: 10\% subset.\label{tab:Ablation-studies}}
\end{table}

\subsection{Ablation Studies}

We conduct ablation studies with our model on CLEVR subset of 10\%
training data (See Table~\ref{tab:Ablation-studies}). We observe
consistent improvements responding to the increase in the number of
reasoning steps as well as going deeper with the representation refinement
process. We have tried up to $p=16$ $\UnitName$ units and $H=16$
GCN layers in each time step, establishing a very deep reasoning process
over hundreds of layers. The results strongly prove the ability to
leverage recurrent cells (row 12-14) and the significance of the deep
refinement layers (row 6-9). It is also clear that linguistic cue
plays a crucial role in all the components of $\ModelName$ and language
binding contributes noticeably to performance (row 1 and 11).

\subsection{Behavior Analysis}

\begin{figure*}
\noindent \begin{centering}
\noindent\begin{minipage}[t]{1\textwidth}%
\begin{center}
\includegraphics[width=0.9\textwidth]{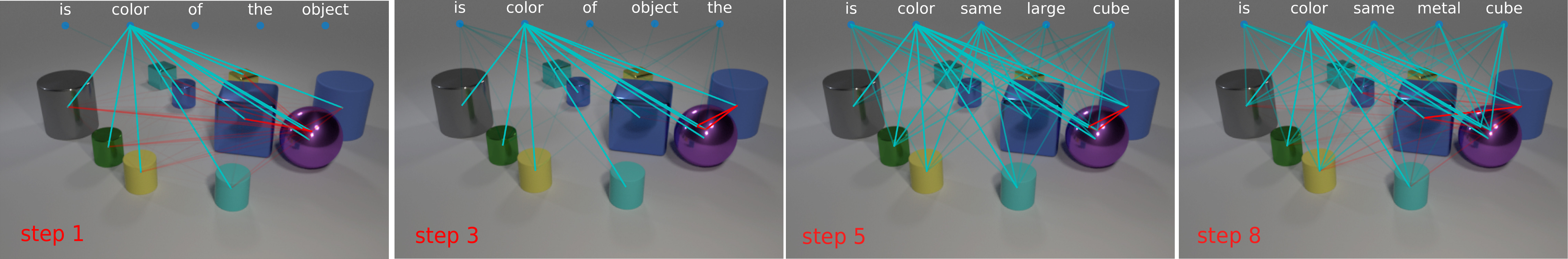}
\par\end{center}
\begin{center}
\begin{minipage}[t]{0.9\columnwidth}%
\textbf{\small{}{}Question}{\small{}{}: Is the color of the big
matte object the same as the large metal cube?}{\small\par}

\textbf{\small{}{}Prediction}{\small{}{}: }\textcolor{green}{\small{}{}yes}\textbf{\small{}{}\qquad{}Answer}{\small{}{}:
yes}{\small\par}%
\end{minipage}
\par\end{center}%
\end{minipage}
\par\end{centering}
\begin{centering}
\ %
\noindent\begin{minipage}[t]{1\textwidth}%
\begin{center}
\includegraphics[width=0.9\textwidth]{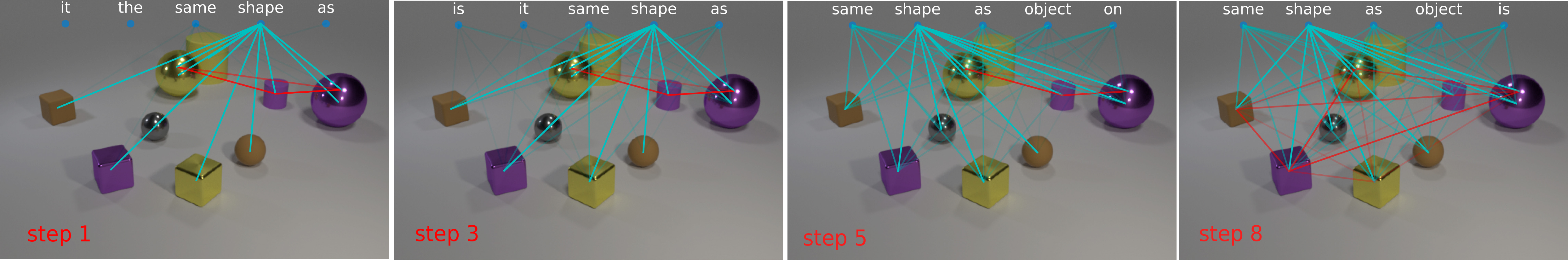}
\par\end{center}
\begin{center}
\begin{minipage}[t]{0.9\columnwidth}%
\textbf{\small{}{}Question}{\small{}{}: There is a tiny purple rubber
thing; does it have the same shape as the brown object that is on
the left side of the rubber sphere?}{\small\par}

\textbf{\small{}{}Prediction}{\small{}{}: }\textcolor{green}{\small{}{}no}\textbf{\small{}{}\qquad{}Answer}{\small{}{}:
no}{\small\par}%
\end{minipage}
\par\end{center}%
\end{minipage}
\par\end{centering}
\noindent \caption{Chains of visual object relation (in red) with language binding (in
cyan) constructed for two image-question pairs. Visual relations are
found adaptively to the specific questions and reasoning stages. Language
binding was sharp on key cross-modality relations at several early
steps, then flats out as memory converges. Only five words included
for visualization purposes. Best viewed in color.\label{fig:qualitative}}
\end{figure*}

To understand the behavior of the dynamic graphs during $\UnitName$
iterations, we visualize them for complex questions from CLEVR (see
Fig.~\ref{fig:qualitative}). As seen, the linguistic objects most
selected for binding are from objects of interest or their attributes
which give a hint to the model of what aspect of the visual cue to
look at. Question types (e.g. yes-no/wh-question, object counting)
and other function words (e.g. ``the'', ``is'', ``on'') are
also paid much attention to. Note that as linguistic objects are outputs
of LSTM passes, those of function words, such as articles and conjunctions
connect nearby content words and holds their aggregated information
through the LSTM operations.

Progressing through the reasoning steps, $\ModelName$ accumulates
multiple aspects of joint domain information in a compositional manner.
In earlier steps when most crucial reasonings happen, it is apparent
in Fig.~\ref{fig:qualitative} that language binding concentrates
on sharp linguistic-visual relations such as from attribute and predicate
words (e.g. ``color'', ``shape'', ``same'') to their related
objects. They constitute the most principal components of the working
memory. Later in the reasoning process, when the memory gets close
to the convergence, the binding weights flat out as not much critical
information is being added anymore. This agrees with the ablation
study result in the last four rows of Table \ref{tab:Ablation-studies}
where the performance raises sharply in the early steps and gradually
converges.

\section{Discussion}

We have presented a new neural recurrent model for compositional and
relational reasoning over a knowledge base with implicit intra- and
inter- modality connections. Distinct from existing neural reasoning
methods, our method computes dynamic dependencies \emph{on-demand}
as reasoning proceeds. Our focus is on $\Problem$ tasks, where raw
visual and linguistic features are given but their relations are unknown.
The experimental results demonstrated superior performance on multiple
datasets even when trained on just 10\% data.

The chaining of implicit relations and representation refinements
in this model suggests further study (a) on the adaptive depth of
refinement layers and the length of the reasoning, e.g., by considering
the complexity of the scene and of the question; and (b) relationship
with first-order logic inference.

 \bibliographystyle{abbrvnat}
\bibliography{LognetIjcai}

\end{document}